\pdfoutput=1

\documentclass[11pt]{article}

\usepackage[]{acl}

\usepackage{times}
\usepackage{latexsym}

\usepackage[T1]{fontenc}

\usepackage[utf8]{inputenc}

\usepackage{microtype}

\usepackage{booktabs}
\usepackage{graphicx}
\usepackage{amsmath}
\usepackage{amssymb}
\usepackage{algorithm}
\usepackage{algorithmic}
\usepackage{enumitem}
\setlist{noitemsep}
\usepackage{colortbl}
\usepackage{subcaption}
\usepackage{arydshln}
\usepackage{xcolor}
\usepackage[normalem]{ulem}

\title{Clip-Tuning: Towards Derivative-free Prompt Learning\\ with a Mixture of Rewards}

 \author{Yekun Chai \, Shuohuan Wang \, \\ \textbf{Yu Sun \, Hao Tian \, Hua Wu \, Haifeng Wang} \\
        Baidu \\ \texttt{\{chaiyekun,wangshuohuan\}@baidu.com}
        \\ \texttt{\{sunyu02,tianhao,wu\_hua,wanghaifeng\}@baidu.com}
        } 

\begin{document}
\maketitle
\begin{abstract}

Derivative-free prompt learning has emerged as a lightweight alternative to prompt tuning, which only requires model inference to optimize the prompts. However, existing work did not take full advantage of the over-parameterized characteristics of large pre-trained language models (PLMs). 
In this paper, we propose Clip-Tuning, a simple yet effective method that adopts diverse frozen ``thinned'' networks of PLMs to obtain \textit{a mixture of rewards} and thus advance the derivative-free prompt learning. The thinned networks consist of all the hidden units that survive a stationary dropout strategy, whose inference predictions reflect an ensemble of partial views over prompted training samples. Our method outperforms previous gradient-free prompt learning methods and achieves parity with gradient-based counterparts on seven language understanding benchmarks under few-shot settings. 





\end{abstract}

\section{Introduction}
\label{sec:intro}




Extensive research has shown that prompt tuning achieves parity with model tuning (\emph{a.k.a.}, full fine-tuning) in few-shot scenarios~\citep{Li2021PrefixTuningOC,Lester2021ThePO}. 
However, prompt tuning that relies on backpropagation from very deep pre-trained transformers requires prohibitive computation and time costs, especially those with billions of parameters~\citep{Brown2020LanguageMA,Rae2021ScalingLM,Wang2021ERNIE3T,Chowdhery2022PaLMSL}. Meanwhile, for many inference-API-based PLMs, researchers either do not have full access to model weights due to commercial restrictions, or cannot afford the training of enormous parameters, which significantly limits the power of derivative-based tuning. Therefore, derivative-free prompt learning has been a promising alternative~\citep{Sun2022BlackBoxTF,Diao2022BlackboxPL}.

%

By treating PLMs as black boxes, derivative-free approaches seem to be a feasible solution to harnessing large PLMs.
\citet{Sun2022BlackBoxTF} leveraged evolutionary algorithms to optimize the continuous prompts by repeatedly calling the inference APIs of PLMs. It adopts the model performance over a spot of samples as the optimization feedback. Nevertheless, few-shot demonstrations can only result in sparse reward, preventing the prompts from taking sufficient informative signals. Hence, our goal is to acquire a mixture of diversified rewards for prompt optimization, using colossal PLMs with millions or billions of parameters in few-shot settings. 

Recent work on \textit{lottery ticket hypothesis}~\citep{Frankle2019TheLT,Chen2020TheLT} states that an over-parameterized PLMs contain matching subnetworks capable of reaching the full test performance comparable to the original model. Then,~\citet{2022DiverseLT, Havasi2021TrainingIS} find that ensembling a mixture of subnetworks can improve the diversity of model predictions and achieve the performance gain. As such, the ensemble of subnetwork predictions is a particularly interesting setting to increase the diversity of learning signals for derivative-free prompt optimization.

Since derivative-free prompt learning only conducts model forward pass without backpropagation, clipping a large proportion of model weights will heavily hurt the overall performance. \citet{Srivastava2014DropoutAS} states that applying dropout to a network amounts to ``sampling'' a thinned network from it. Moreover, previous work~\cite{Gao2021SimCSESC,DBLP:conf/nips/LiangWLWMQCZL21} finds that standard dropout can 
act as ``minimal data augmentation'' to construct different sample representations. Therefore, we employ dropout during model inference to ``sample'' different ``thinned'' subnetworks and diversify the data representations. Note that our subnetworks are deterministic, while the original dropout is random at each time. For each training example, diverse subnetworks produce a variety of hidden representations and fitness rewards, which diversifies the learning feedback for gradient-free prompt learning.

\paragraph{Contributions} (1) We propose a simple yet effective method Clip-Tuning, in which multiple frozen subnetworks act as multi-view critics and provide a mixture of informative rewards for gradient-free prompt optimization (\S\ref{sec:clip}). The importance and originality of this study are that it explores the new direction of the exploitation of \textit{reward diversity} in gradient-free prompt optimization. (2) Empirical results show that our method has surpassed previous gradient-free prompt learning approaches on seven natural language understanding (NLU) benchmarks in few-shot settings (\S\ref{sec:exp}). Surprisingly, the \emph{random search} method can serve as an excellent few-shot baseline to prime large PLMs. 
(3) Our method sheds light on inference-only PLMs and can be a good fit for commercial PLM providers to build API-based features. Note that our method requires API providers to support dropout operation, whereas API users do not need to make any change based on derivative-free prompt learning.




\section{Related work}
\label{sec:bg}

\subsection{Prompt-based learning}
Holding the promise of exploiting the few-shot learning capability of large pre-trained models, prompt-based learning has attracted extensive attention in recent years~\citep{Brown2020LanguageMA,Schick2021ExploitingCF,Li2021PrefixTuningOC,Lester2021ThePO,Sun2022BlackBoxTF}. It primes the frozen PLMs using a series of discrete natural language tokens or continuous ``soft prompts'' to conduct various downstream tasks. Early work employed exemplar language templates to condition the PLMs for task-specific prediction~\citep{schick2021few, scao2021many}. Such methods require manual involvement of humans in the design of prompt templates, making the continuous prompt a promising direction.

\paragraph{Prompt tuning} Prompt tuning approaches~\cite{Li2021PrefixTuningOC, Lester2021ThePO,liu2021gpt} prepend a string of continuous word embeddings as ``virtual tokens'' to prime the pre-trained models, where it optimizes the continuous prompts with backpropagation while freezing the model weights of PLMs. These methods achieve parity with full model tuning and even surpasses the fine-tuning in few-shot settings.

\paragraph{Prompt search} There has been a surge of interest in automatic prompt learning, which treats the prompt as a parameter space to be optimized over. One line of prompt search methods focuses on \textit{discrete} search space, \emph{i.e.}, natural language tokens. \citet{shin2020autoprompt} employ a gradient-based method to find the optimal trigger words to construct the prompt. \citet{prasad2022grips} use a gradient-free edit-based search method to refine the instructional language prompts, in which it produces the optimal edited prompts given manually designed ones. Another line in this direction is \textit{continuous} prompt search, where the prompt is optimized as ``virtual tokens'' in the continuous parameter space. \citet{Sun2022BlackBoxTF} adopt the Covariance Matrix Adaptation Evolutionary Strategy (CMA-ES)~\citep{Hansen2003ReducingTT} to search over the intrinsic dimension of prompts~\citep{Aghajanyan2021IntrinsicDE} with only access to the inference API of PLMs. This approach only requires the forward pass of PLMs without the need for gradient backpropagation. This work builds upon this line of research, targeting better exploiting the over-parameterization of PLMs to collect fine-grained rewards for search algorithms.


\subsection{Derivative-free optimization}  Derivative-free optimization targets the settings that the derivative of the objective is unavailable or unreliable to achieve. It iteratively optimizes the parameter candidate by local hill-climbing in the objective landscape. Suppose the objective function $f: A \rightarrow \mathbb{R}$ for some set $A$, derivative-free optimization only uses the input $x$ and its fitness $f(x)$ after evaluation for iterative optimization. Examples include evolutionary strategies~\citep{Hansen2003ReducingTT},  Bayesian optimization~\citep{frazier2018tutorial}, random search~\citep{zabinsky2009random}, and so forth. In this work, we experimented with CMA-ES~\citep{Hansen2003ReducingTT} and pure random search~\citep{zabinsky2009random} algorithms.

\section{Derivative-free prompt learning}
\label{sec:pre}
Vanilla derivative-free prompt learning~\citep{Sun2022BlackBoxTF} employs the model inference to evaluate the fitness of candidate prompts for iterative prompt learning using evolutionary algorithms. Firstly, it prepends a series of soft prompt embeddings $P$ to the input tokens $X$ to feed the prompted input $[P;X]$ into the frozen pre-trained transformers $f$ parameterized by $\theta$. The prompt $P = P_0 + P_\Delta$ is the summation of randomly initialized or pre-trained prompt $P_0 \in \mathbb{R}^D$ and prompt change $P_\Delta \in \mathbb{R}^{D}$ that is iteratively optimized by the Covariance Matrix Adaptation Evolutionary Strategy (CMA-ES)~\citep{Hansen2003ReducingTT}. Meanwhile, \citet{Aghajanyan2021IntrinsicDE} find that PLMs have a low dimension reparameterization. The search space of $P_\Delta$ can be reparameterized as a intrinsic dimensionality $\mathbf{z} \in \mathbb{R}^d ~(d \ll D)$ with a randomly initialized affine transformation $\mathbf{W}\in\mathbb{R}^{d \times D}$; that is, $P_\Delta = \mathbf{z} \cdot \mathbf{W}$. Thus, the prompt is optimized over an intrinsic dimensionality 
of $\mathbf{z}$ instead of $P_\Delta$. Then it performs task-specific inference to get the fitness of candidate prompts using an arbitrary objective function $\mathcal{L}(f_{\theta}([P;X]), y)$, where $y$ denotes the golden standard label for the input $X$. Afterward, the CMA-ES algorithm iteratively optimizes the prompt fitness to get the optimum prompt that satisfies $\arg\min_{P} \mathcal{L} (f_{\theta}([P;X]), y)$. 


\section{Methodology}
\label{sec:method}

We propose Clip-Tuning by adopting multiple deterministic clipping instances of PLMs to form diverse views of critics on candidate soft prompts. 
It has two key properties: (i) Only an instance of frozen model weights is used to acquire diverse reward signals (\emph{i.e.}, fitness) of candidate prompts. Additional budgets include the storage of clipping dropout masks. (ii) Users only require the model forward pass via API without knowing the model internals during prompt learning.

\begin{figure}[h]
\vskip -2mm
\begin{center}
\includegraphics[width=\columnwidth]{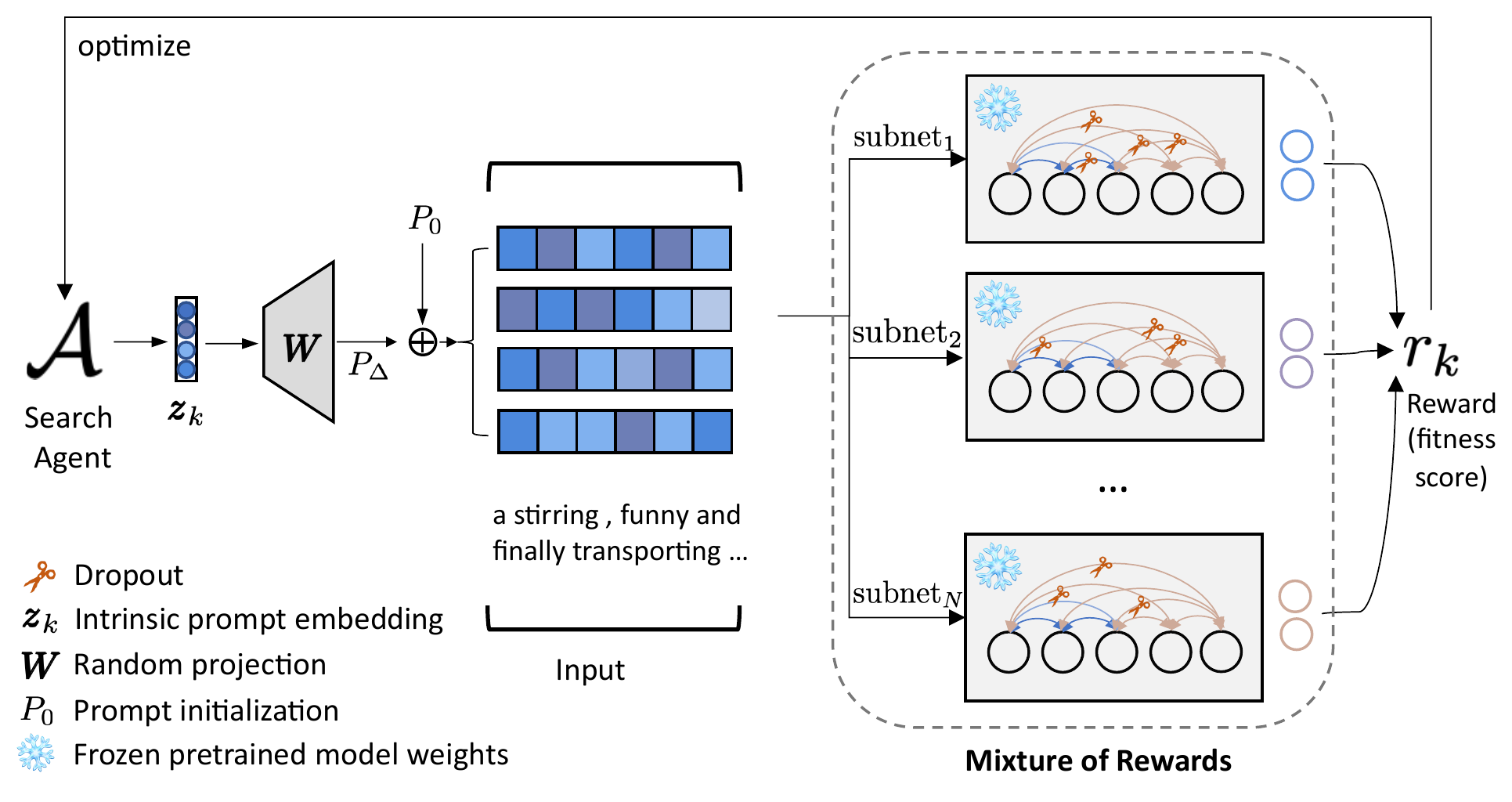}
\vskip -1mm
\caption{Illustration of Clip-Tuning process.}
\label{fig:DLT}
\end{center}
\vskip -3mm
\end{figure} 


\subsection{Dropout as Language Model Clip}
\label{sec:clip}

As shown in Figure~\ref{fig:DLT} (\dashuline{dashed area}), Clip-Tuning employs the dropout clipping multiple ($N$) times to get a suite of subnetworks of frozen PLMs. As aforementioned, we only apply the clipping \textit{in the place of dropout layers during training} to get the subnetworks with minimal costs, making the inference settings the same as training. 

For each clipping layer in the $j$-th subnetwork ($j=1,2,\cdots,N$), it generates the binary mask $\tilde{\mathbf{M}}_j$ drawn from a Bernoulli distribution with a ratio of probability $p_\textrm{clip}$ and a unique seed $s_j$ generated by a string hash algorithm, \emph{i.e.}, SHA-1 hash function~\citep{burrows1995secure}, to keep each instance's clipping mask fixed (see Appendix \S~\ref{sec:code} for the pseudocode):
\begin{align}
    \mathbf{M}_j \sim \textrm{Bernoulli} (p_\textrm{clip}, s_j) \label{eq:eq1}
\end{align}{As a result, we produce the static binary mask $\tilde{\mathbf{M}}_j$ for target clipping layers of each subnetwork to keep it unique. In this case, each unit is removed with a fixed probability $p_\textrm{clip}$ independent of other units, where $p_\textrm{clip}$ is chosen by the validation set.
}


After removing a proportion of internal hidden units, the output of each clipped layer is zoomed out to approximately $(1-p_\textrm{clip})$ of the counterpart of original PLMs. Following the inverted dropout operation~\citep{Srivastava2014DropoutAS}, we apply the rescaling by dividing the survived units of clipped layers in each subnetwork by a factor of $(1-p_\textrm{clip})$ to keep the output expectation unchanged.

Therefore, given the original units $\mathbf{o}$ and $j$-th clipping mask $\mathbf{M}_j$, the resultant units can be :
\begin{align} \label{eq:eq2}
    \mathbf{o}^{'}_j = \frac{\mathbf{o}}{1-p_\textrm{clip}} \odot \mathbf{M}_j
\end{align}{where the output units should be rescaled to as per normal to make an inference.}

\subsection{Prompt learning with Clip-Tuning}
As shown in Algorithm~\ref{alg:1}, Clip-Tuning renders diverse predictions from a population of subnetworks, which enhances the diversity of prompt learning rewards. For each subnetwork, we utilize a randomly initialized linear layer on top of ``[CLS]'' embeddings for class prediction. Given the subnetwork prediction $f_\theta(\cdot)$ and golden labels $y$, we then use the evaluation function $\mathcal{L}$ to compute the fitness score for each subnetwork. With multiple subnetworks, the overall prompt fitness $r_k$ is computed as the average from different subnetworks for the use of prompt optimization. The prompt learning agent $\mathcal{A}$ can take various derivative-free algorithms. Following~\citet{Sun2022BlackBoxTF}, we use the CMA-ES algorithm~\citep{Hansen2003ReducingTT} for prompt learning and leave others for future work.

\begin{algorithm}[]
\begin{algorithmic}[1]
  \STATE {\bfseries Require:} Clipping mask set $\mathcal{M} =\{m_1, m_2,..., m_N \}$; frozen subnetwork tribe $\{f_{\theta} (\cdot \vert m) \vert \forall m \in \mathcal{M} \}$; few-shot samples $X$ and labels $y$; the fitness evaluation function $\mathcal{L}(\cdot)$; the maximum update steps $K$, initialized prompt embedding $P_0 \in \mathbb{R}^{D}$, randomly initialized affine transformation $\mathbf{W}\in \mathbb{R}^{d\times D}$,  search agent $\mathcal{A}$ with derivative-free algorithms.
    \FOR{$k=\{1,2,\cdots, K\}$ steps}
    \STATE Sample an intrinsic embedding of candidate prompt $\mathbf{z}_k \in \mathbb{R}^{d}$ from the agent $\mathcal{A}$;
    \STATE Get prompt embeddings: \\
    $\displaystyle
       \qquad    P_k = P_0 +  P_\Delta = P_0 + \mathbf{z}_k \cdot \mathbf{W}
    $
    \STATE Evaluate the fitness $r_k$ of candidate prompts $P_k$ using subnetworks  \colorbox{blue!10}{$\mathbf{\rhd}$ \textbf{Clip-Tuning}} \\
    $\displaystyle
       \qquad    r_k = \mathbb{E}_{m \sim \mathcal{M}}\mathcal{L} \big(f_{\theta}([P_k;X]| m), y \big)
    $
    \STATE Optimize the search agent $\mathcal{A}$ using $r_k$.
    \ENDFOR
\end{algorithmic} 
  \caption{Derivative-free prompt learning with Clip-Tuning.}
  \label{alg:1}
\end{algorithm}


\section{Experiments and results}
\label{sec:exp}
\paragraph{Datasets and evaluation metrics} We conduct experiments on 
seven language understanding datasets, including SST-2~\citep{Socher2013RecursiveDM}, Yelp polarity, AG's News, DBPedia~\citep{Zhang2015CharacterlevelCN}, SNLI~\citep{Bowman2015ALA}, RTE~\citep{Wang2018GLUEAM}, and MRPC~\citep{Dolan2005AutomaticallyCA}. We randomly draw 16 samples for each class as training and evaluation sets, respectively, to construct true few-shot learning. All datasets except MRPC adopt accuracy for evaluation, whereas MRPC uses F1-measure. Appendix~\S\ref{sec:data} reports the detailed data statistics.

\paragraph{Models and optimization}\,\,
 We follow the same baseline setting as~\citet{Sun2022BlackBoxTF} for few-shot learning. We utilize RoBERTa\textsubscript{Large}~\citep{Liu2019RoBERTaAR} in all experiments. \textit{Derivative-based methods} include  prompt tuning~\citep{Lester2021ThePO}, adapter tuning~\citep{houlsby2019parameter}, Low-Rank Adaptation (LoRA;~\citealp{Hu2022LoRALA}), full model tuning, and feature-based tuning~\citep{Peters2019ToTO} that uses a trainable bi-LSTM or linear classifier on the top for prediction. \textit{Derivative-free methods} consist of a manual prompt that uses manually designed templates for zero-shot evaluation; in-context learning~\citep{Brown2020LanguageMA}; random search that randomly samples a large set of search points then selects the optimal after evaluation; and Black-Box Tuning~\citep{Sun2022BlackBoxTF} that harnesses the CMA-ES algorithm for derivative-free search. \textit{Our method} takes the hinge loss as $\mathcal{L}$ for reward computation, and a clipping ratio of 0.1. Following~\citet{Sun2022BlackBoxTF}, we randomly draw tokens from RoBERTa vocabularies as prompt initialization $P_0$ for text classification tasks while adopting prompt embedding pretrained on MNLI~\citep{Williams2018ABC} for two-sentence tasks. Appendix~\S\ref{sec:impl} lists the optimization details.
\begin{table*}[t]
\resizebox{\textwidth}{!}{%
\begin{tabular}{lcllllllll}
\hline
 & \textbf{\begin{tabular}[c]{@{}c@{}}Tunable\\ params\end{tabular}} & \textbf{SST-2} & \textbf{Yelp Polarity} & \textbf{AG's News} & \textbf{DBPedia} & \textbf{MRPC} & \textbf{SNLI} & \textbf{RTE} & \textbf{Avg.} \\ \hline
\multicolumn{10}{c}{Discrete prompts} \\ \hline
Manual Prompt & 0 & 79.82 & 89.65 & 76.96 & 41.33 & 67.40 & 31.11 & 51.62 & 62.56 \\
In-Context Learning & 0 & 79.79{\small $\pm$3.06} & 85.38{\small $\pm$3.92} & 62.21{\small $\pm$13.46 } & 34.83{\small $\pm$7.59} & 45.81{\small $\pm$6.67} & 47.11{\small$\pm$0.63} & 60.36{\small $\pm$1.56} & 59.36 \\ \hline
\multicolumn{10}{c}{Derivative-based optimization} \\ \hline
(Full) Model Tuning & 355M & 85.39{\small $\pm$2.84} & 91.82{\small $\pm$0.79} & \textbf{86.36}{\small$\pm$1.85} & \begin{tabular}[c]{@{}l@{}}97.98{\small $\pm$0.14}\end{tabular} & \begin{tabular}[c]{@{}l@{}}77.35{\small $\pm$5.70}\end{tabular} & \begin{tabular}[c]{@{}l@{}}54.64{\small $\pm$5.29}\end{tabular} & 58.60{\small $\pm$6.21} & 78.88 \\
Feature-based (MLP) & 1M & 64.80{\small $\pm$1.78} & 79.20{\small $\pm$2.26} & 70.77{\small $\pm$0.67} & 87.78{\small $\pm$0.61} & 68.40{\small $\pm$0.86} & 42.01{\small $\pm$0.33} & 53.43{\small $\pm$1.57} & 66.63 \\
Feature-based (LSTM) & 17M & 65.95{\small $\pm$0.99 } & 74.68{\small $\pm$0.10} & 77.28{\small $\pm$2.83} & 90.37{\small $\pm$3.10} & 71.55{\small $\pm$7.10} & 46.02{\small $\pm$0.38} & 52.17{\small $\pm$0.25} & 68.29 \\
Prompt Tuning & 50K & 68.23{\small $\pm$3.78} & 61.02{\small$\pm$6.65} & 84.81{\small $\pm$0.66} & 87.75{\small $\pm$1.48} & 51.61{\small $\pm$8.67} & 36.13{\small $\pm$1.51} & 54.69{\small $\pm$3.79} & 64.07 \\
\quad w/ pre-trained prompt & 50K & - & - & - & - & 77.48{\small $\pm$4.85} & 64.55{\small $\pm$2.43} & 77.13{\small $\pm$0.83} & 74.42 \\
Adapter Tuning & 2.4M & 83.91{\small $\pm$2.90} & 90.99{\small $\pm$2.86} & 86.01{\small $\pm$2.18} & \textbf{97.99}{\small $\pm$0.07} & 69.20{\small $\pm$3.58} & 57.46{\small $\pm$6.63} & 48.62{\small $\pm$4.74} & 76.31 \\
LoRA & 786K & 88.49{\small $\pm$2.90} & 90.21{\small $\pm$4.00} & 87.09{\small $\pm$0.85} & 97.86{\small $\pm$0.17} & 72.14{\small $\pm$2.23} & 61.03{\small $\pm$8.55} & 49.22{\small $\pm$5.12} & 78.01 \\
\hline
\multicolumn{10}{c}{Derivative-free optimization} \\ \hline
Black-Box Tuning & 500 & 89.34{\small $\pm$1.07} & 91.26{\small $\pm$0.40} & 82.04{\small $\pm$0.63} & 79.43{\small $\pm$0.02} & 61.56{\small $\pm$4.34} & 46.58{\small $\pm$1.33} & 52.59{\small $\pm$2.21} & 71.83 \\
\quad w/ pre-trained prompt & 500 & - & - & - & - & 75.58{\small $\pm$1.63} & 81.02{\small $\pm$1.39} & 77.62{\small $\pm$1.65} & 82.33 \\
Random Search & 500 & 86.35 & 89.92 & 79.96 & 37.15 & 76.03 & 40.08 & 51.63 & 65.87 \\\hdashline
Clip-Tuning (ours) & 500 & \textbf{90.44}{\small $\pm$0.98} & \textbf{92.15}{\small$\pm$0.43} & 82.81{\small $\pm$0.12} & {80.38\small$\pm$0.03} & 75.12 & 50.62 & 53.43 & 74.99 \\
\quad w/ pre-trained prompt & 500 & - & - & - & - & \textbf{77.78}{\small$\pm$1.22} & \textbf{83.50}{\small $\pm$1.38} & \textbf{78.25}{\small$\pm$1.54} & \textbf{83.62} \\ \hline
\end{tabular}%
}
\caption{Overall \textit{16-shot} performance of gradient-based or gradient-free methods on seven NLU benchmarks.}
\label{tab:results}
\end{table*}

\paragraph{Main results} Table~\ref{tab:results} compares the test-set performance of baselines in few-shot settings\footnote{Note that the reproduced DBPedia results of Black-Box Tuning are different from~\citep{Sun2022BlackBoxTF}. We conjecture it is due to experimental factors such as random seeds.}. Our method outperforms the previous prompt learning method (Black-Box Tuning) on seven benchmarks. The improvement on three sentence-pair classification datasets (+1.77\% on average) outperforms the counterpart on four text classification datasets (+0.93\%). It is observed that our method confers amenable benefits using both randomly initialized prompts and pre-trained prompts, which achieves more considerable performance gain using randomly initialized prompts. 
Interestingly, the \textit{pure random search} can serve as a good baseline for few-shot comparison, even achieving competitive results on MRPC.

\begin{table}[thb]
\vskip -2mm
\centering
\resizebox{\columnwidth}{!}{%
\begin{tabular}{|cccccc|}
\hline
Dataset & Task & \#class & w/ Clip-Tuning & w/o Clip-Tuning & $\Delta$perf \\ \hline
SST-2 & SA & 2 & 90.14 & 88.65 & -1.49 \\
Yelp Polarity & SA & 2 & 92.36 & 91.16 & -1.20 \\
AG's News & TC & \cellcolor{lightgray}4 & 82.66 & 81.72 & -0.94 \\
DBPedia & TC & \cellcolor{lightgray}14 & 80.80 & 79.43 & -1.34 \\\hdashline
MRPC & PC & 2 & 78.69 & 76.56 & \cellcolor{lightgray}-2.09 \\ 
SNLI & NLI & 3 & 83.16 & 80.84 & \cellcolor{lightgray}-2.32 \\
RTE & NLI & 2 & 79.78 & 75.45 & \cellcolor{gray}-4.33 \\ \hline
\end{tabular}%
}
\caption{Ablation test performance. SA: sentiment analysis; TC: topic classification; PC: paraphrase identification; NLI: natural language inference.}
\label{tab:ablation}
\vskip -2mm
\end{table}

\paragraph{Ablation study} To examine the benefits of proposed method, we conduct the ablation test by removing the Clip-Tuning, as shown in Table~\ref{tab:ablation}. The ablation test results suggest that our method provides more informative feedback for sentence-pair tasks than single-sentence classification. Moreover, the performance gains grow when the number of categories becomes small.



\begin{figure}[thb]
\vskip -1mm
\centering
\includegraphics[width=\columnwidth]{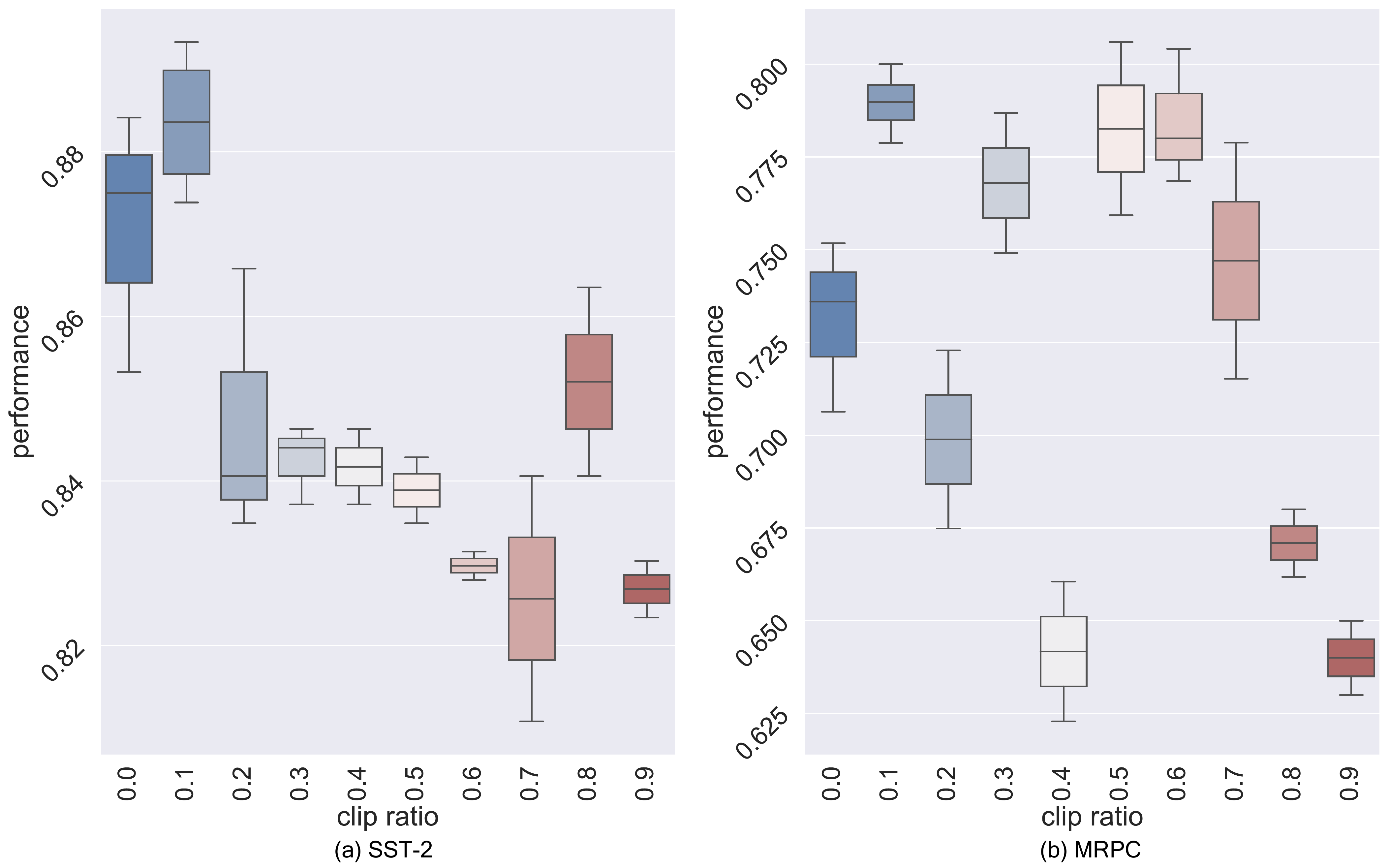}
\caption{Varying clip ratios vs. performance.}
\label{fig:pdrop}
\vskip -2mm
\end{figure}

\paragraph{Varying clip ratios} Figure~\ref{fig:pdrop} illustrates the test-set performance versus various clip ratios. The performance on SST-2 peaks at the clipping ratio  ($p_\text{clip}$) of 0.1, which is the same as the RoBERTa dropout rate. The performance curve of MRPC shows a saddle trend: it first peaks at 0.1 and then rebounds to the top at 0.5. We conjecture that the pre-trained model weights fit in with the dropout during pre-training, thereby performing smoothly with the identical clip rate.  


\begin{figure}[h]
\begin{center}
\includegraphics[width=\columnwidth]{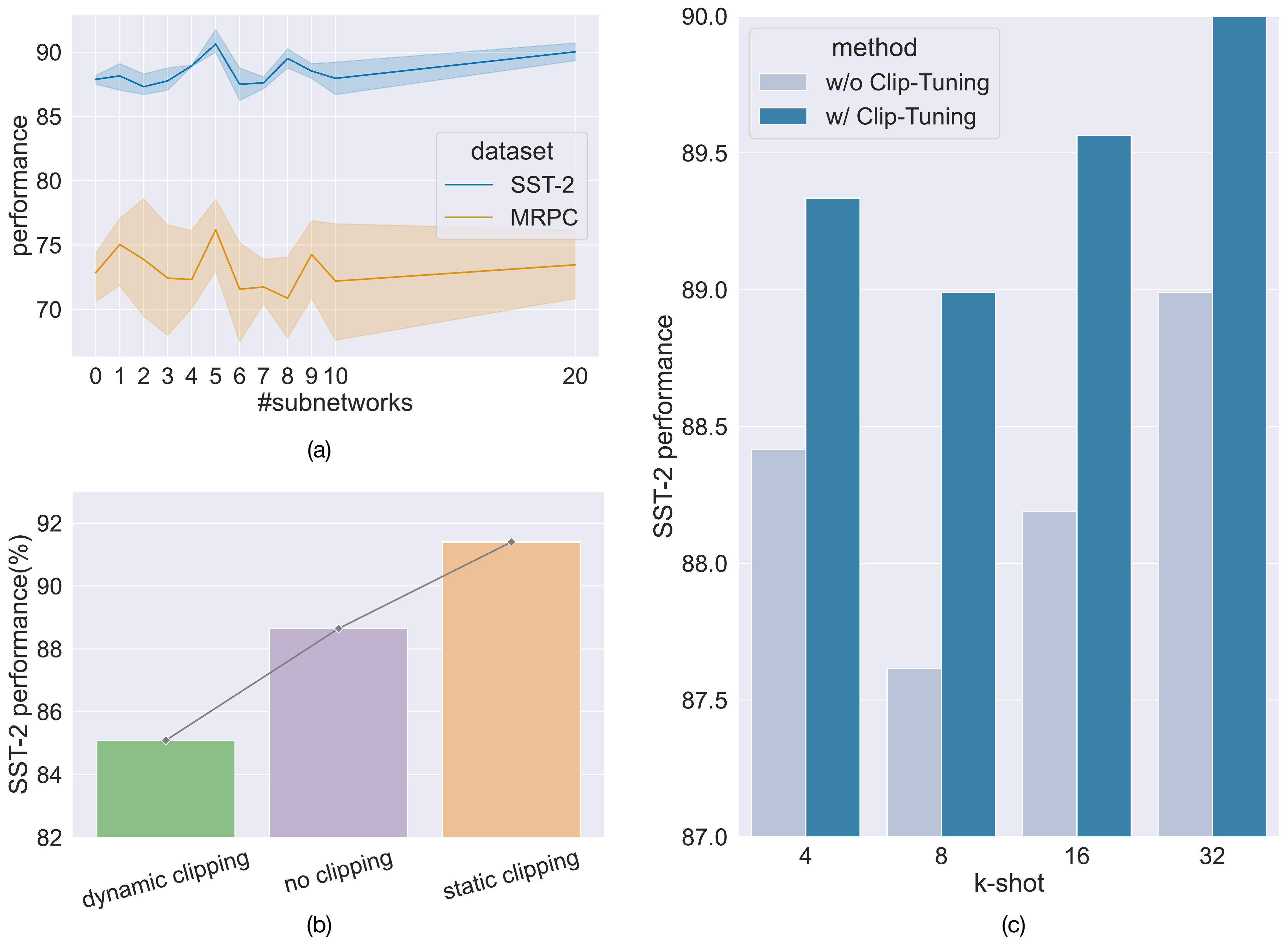}
\caption{(a) Varying subnetworks ($N$) vs. performance. (b) Dynamic / no / static clipping vs. performance. (c) Size of training samples (SST-2).}
\label{fig:comb}
\end{center}
\vskip -3mm
\end{figure}

\paragraph{Varying thinned subnetworks ($N$)} Figure~\ref{fig:comb}(a) reveals that there has been a fluctuated increment as the number of thinned subnetworks increases. More subnetworks can deliver more fine-grained learning rewards but may be impeded by the numerous inference costs. We thus take the first peak ($N \approx 5$) for economic consideration.

\paragraph{Static vs. dynamic clipping} Our method applies the static clipping for each thinned subnetwork. Figure~\ref{fig:comb}(b) compares the dynamic clipping with random dropout masks every time, static clipping (ours), and no clipping on SST-2. It is obvious that static clipping confers benefits whereas dynamic clipping hurts. This is because dynamic clipping results in unexpected noise and thereby misdirecting the search direction during prompt learning.

\paragraph{Size of training data} Figure~\ref{fig:comb}(c) exhibits the effect of our method on various training data sizes: Clip-Tuning consistently improves upon the original soft prompts. There has been a generally stable performance gain on prompt learning after applying the Clip-Tuning approach. 

\paragraph{Discussion} Our method improves from the API provider side, which requires trivial modifications of API services (\emph{i.e.}, supporting dropout), while it still remains a black box for API users. It can be easily scaled for deployment in a distributed manner. Each serving PLMs only needs to support the dropout operation to form the subnetwork and return the score over few-shot samples.





\section{Conclusion and future work}
\label{sec:concl}
We propose Clip-Tuning, a simple method to enhance derivative-free prompt learning for NLU tasks. It focuses on acquiring fine-grained informative feedback for derivative-free optimization, beating the previous rivals on seven NLU benchmarks. Our method sheds light on the exploration of inference-only API-based PLMs. In the future, it is a promising direction to reduce the cost of green derivative-free prompt optimization.





\section*{Limitations}
\label{sec:limit}
    Although our method optimizes over a small size ($d=500$) of intrinsic dimensionality, it evaluates the fitness of training samples with multiple subnetworks, costing extra computation resources. Besides, derivative-free prompt learning is constrained by the limited intrinsic dimensionality, grounding on the findings of large PLMs~\citep{Aghajanyan2021IntrinsicDE}. Further, we empirically find it more suitable for NLU tasks with a small class number. However, it is still a good fit for exploiting the API-based tuning by only applying the model inference. 





\section*{Acknowledgements}
The authors thank all anonymous reviewers for
their insightful and constructive comments. 


\bibliography{anthology,custom}
\bibliographystyle{acl_natbib}

\clearpage

\appendix
\section{Appendices}

\subsection{Pesudocode}
\label{sec:code}

Algorithm~\ref{alg:2} presents the pseudocode to get fixed subnetworks. It is simple to apply the clipping operation in the place of dropout layer in PLMs with few lines.
\begin{algorithm}[]
\begin{algorithmic}[1]
    \STATE Define a unique string and convert to unique integer seed $s$ using string hash algorithm, \emph{i.e.}, SHA-1 algorithm, for each clipping layer;
    \STATE Generate a unique clipping mask according to Eq.~\eqref{eq:eq1}; 
    \STATE Apply the clipping operation as Eq.~\eqref{eq:eq2} to obtain the unique ``thinned'' subnetwork;
    \STATE Store the unique string $s$ (rather than subnetwork weights) for reuse and memory-saving.
\end{algorithmic} 
  \caption{Fixed clipping to get a single deterministic subnetwork.}
  \label{alg:2}
\end{algorithm}





        

\subsection{Data statistics}
\label{sec:data}
Table~\ref{tab:data_stat} summarizes the statistics of natural language understanding
benchmark datasets for evaluation. We randomly sample 16 samples for the training and evaluation set to construct the true few-shot training and evaluation set, respectively. Following~\citep{Zhang2021RevisitingFB,Gao2021MakingPL, Gu2021PPTPP, Sun2022BlackBoxTF}, we employ original evaluation set as the test set. We use the test set for those without a development set to evaluate the performance.

\begin{table}[thb]
\resizebox{\linewidth}{!}{%
\begin{tabular}{@{}llrrr@{}}
\toprule
Data          & NLU Task                      & \#train & \#test & \#class \\ \midrule
SST-2         & Sentiment analysis        & 67k     & 0.9k   & 2       \\
Yelp polarity & Sentiment analysis        & 560k    & 38k    & 2       \\
AG's News     & Topic classification      & 120k    & 7.6k   & 4       \\
DBPedia       & Topic classification      & 560k    & 70k    & 14      \\
MRPC          & Paraphrase classification & 3.7k    & 0.4k   & 2       \\
SNLI          & Language inference        & 549k    & 9.8k   & 3       \\
RTE           & Language inference        & 2.5k    & 0.3k   & 2       \\ \bottomrule
\end{tabular}%
}
\caption{Summary of data statistics.}
\label{tab:data_stat}
\end{table}

\subsection{Training details}
\label{sec:impl}

We summarize the training settings of baseline models as follows: (1) Prompt Tuning. Following~\citep{Lester2021ThePO}, only the soft prompts prepended to the input sequences are trained while freezing the PLM model weights. The Adam optimizer with a learning rate of 5e-4, and a batch size of 16 for 1,000 epochs are adopted. (2) Model Tuning. It uses the Adam optimizer with a learning rate of 1e-5 and batch size 16 for 200 epochs. (3) Feature-based Tuning~\citep{Peters2019ToTO}. The weights of PLMs are frozen, and only the BiLSTM or MLP layer on top of {\verb |[CLS]| } are trained. Both settings apply the Adam optimizer with a learning rate of 3e-4, batch size of 16, and 1,000 training epochs on few-shot data. (4) Adapter tuning~\citep{houlsby2019parameter}. The weights of PLMs are frozen, Adapter injects the parameter-efficient adapter module after each transformer sub-layer, but before the skip connection. In our experiment with RoBERTa\textsubscript{large}, it only tunes around 2.4M extra parameters in total. (5) Manual prompt. The following four manually designed prompts are used for zero-shot evaluation: 
\begin{enumerate}
    \item  ``<sentence>. It was [MASK].''
    \item  ``[MASK] news: <sentence>''
    \item  ``[Category: [MASK]] <sentence>''
    \item  ``<hypothesis>? [MASK], <premise>''
\end{enumerate}{In which SST-2 and Yelp Polarity datasets adopt template 1, AG's News uses template 2, DBPedia uses template 3, and all two-sentence classification datasets including MRPC, RTE, and SNLI apply template 4.
}

(6) In-context learning~\citep{Brown2020LanguageMA}. Randomly selected 32 training samples are concatenated with the input text for GPT-3 like in-context learning. (7) Random Search. We apply the random search method on the intrinsic prompt embedding $\mathbf{z}_k \in \mathbb{R}^d (d=500)$  for 1,000 steps. It samples a large number of search points and selects the optimal one based on their fitness. (8) Black-Box Tuning. We follow the original experimental settings in~\citep{Sun2022BlackBoxTF} which adopt the CMA-ES algorithm for black-box search.

For a fair comparison, all prompt-based methods apply the same prompt length, manual template, and initialized prompt embeddings $P_0$. All experiments are run over multiple random seeds on 4 Nvidia A100 chips. It is worth noting that different subnetworks will not retard the training time since we parallelly evaluate different subnetworks.

For Clip-Tuning methods, we adopt the following hyperparameters setting as shown in Table~\ref{tab:hyperparams}. We adopt the same CMA-ES algorithm settings in~\citet{Sun2022BlackBoxTF}, \emph{i.e.}, intrinsic prompt dimension $d=500$, evolutionary population size of 20. The margin of hinge loss $\mathcal{L}$ is set to 2.

\begin{table}[thb]
\resizebox{\columnwidth}{!}{%
\begin{tabular}{@{}ll@{}}
\toprule
Hyperparameter                   & Value                               \\ \midrule
Intrinsic dimension ($|z_k|$)    & 500                                 \\
Prompt length                & 50                                  \\
Clipping ratio ($p_\textrm{clip}$)      & 0.1                        \\
Number of subnetworks ($N$)   & 5                                   \\
Search algorithm ($\mathcal{A}$) & CMA-ES  \\
Search iteration $k$             & 10k, except 20k for DBPedia \\ \bottomrule
\end{tabular}%
}
\caption{Hyperparameter settings of our experiments.}
\label{tab:hyperparams}
\end{table}

\subsection{Choices of fitness evaluation function $\mathcal{L}$}
To verify the effect of the fitness evaluation function, we design experiments with a cross-entropy function on four types of NLU tasks, one dataset for each task (see Table~\ref{tab:data_stat} for statistics). Table~\ref{tab:loss_func} shows that Clip-Tuning can also achieve performance gains with cross-entropy functions on SNLI tasks but degrades on other three datasets. However, the improvement consistency is worse than hinge loss. This may be because that hinge loss is less prone to outliers in training samples, making the optimization process more robust.

\begin{table}[thb]
\resizebox{\columnwidth}{!}{%
\begin{tabular}{|cllll|}
\hline
Fitness function $\mathcal{L}$ & SST-2 & AG's News & MRPC  & SNLI  \\ \hline
ce                             & 88.87 & 83.49     & 73.72 & 80.38 \\
ce + ours            & 88.20 & 82.82     & 70.60 & 83.08$\uparrow$\\ \hdashline
hinge                          & 87.04 & 81.73     & 72.18 & 80.84 \\
hinge+ours           & 90.44$\uparrow$ & 82.81$\uparrow$     & 72.46$\uparrow$ & 83.16$\uparrow$ \\ \hline
\end{tabular}%
}
\caption{Fitness evaluation function comparison. Cross-entropy (ce) vs. hinge loss. Experiments are run over multiple random seeds.}
\label{tab:loss_func}
\vskip -3mm
\end{table}



\subsection{Random vs. pre-trained prompt initialization ($P_0$)} 
\begin{figure}[!h]
\begin{center}
\includegraphics[width=\columnwidth]{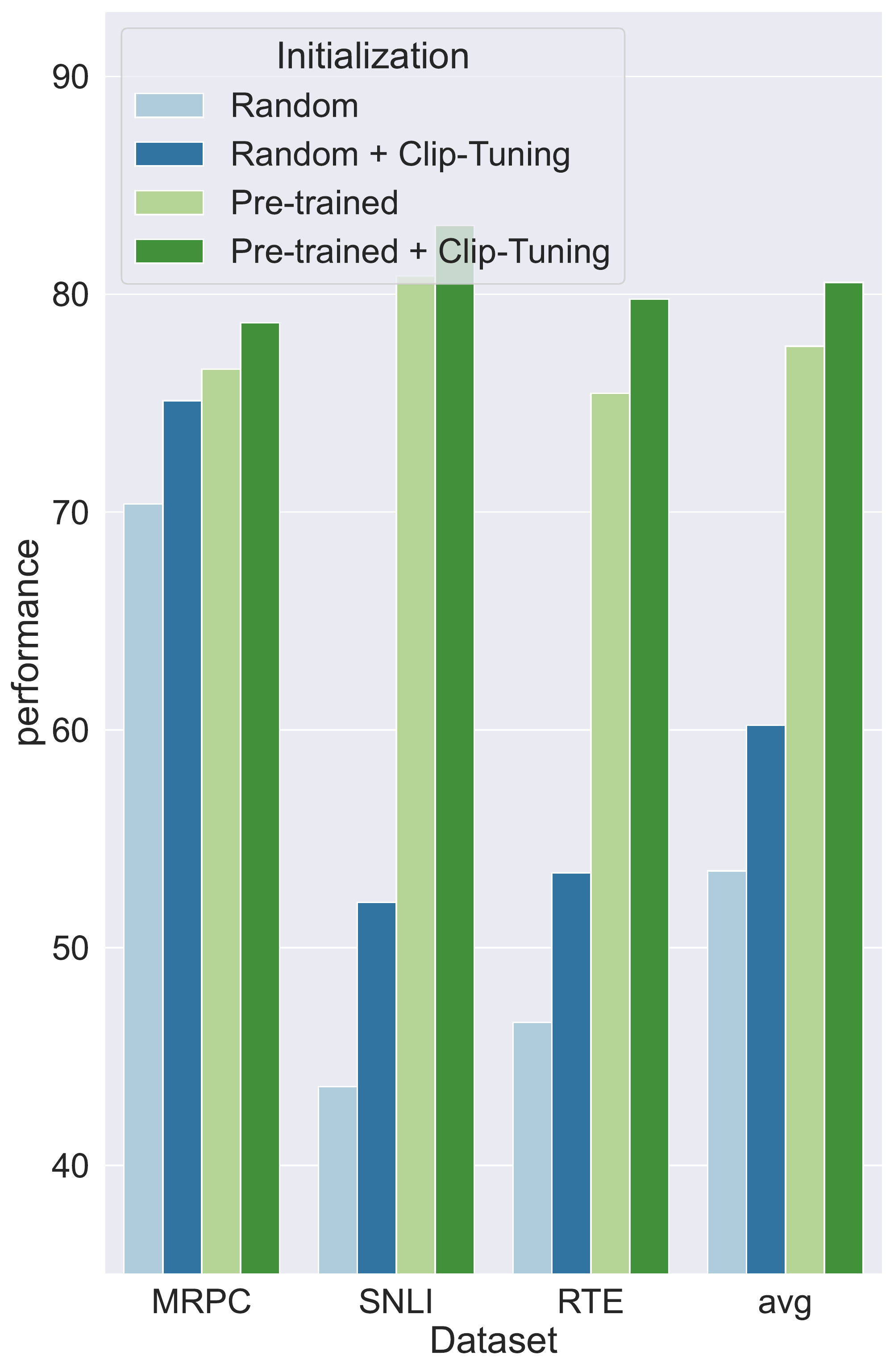}
\caption{Different prompt initialization vs. performance (SST-2).}
\label{fig:prompt_init}
\end{center}
\end{figure}

To testify the generalized effect of our method using various prompt initialization on sentence-pair tasks, we conduct further experiments as shown in Figure~\ref{fig:prompt_init}. It is observed that our method confers amenable benefits using both initialization strategies on three benchmarks. Meanwhile, our method achieves more considerable performance gain using randomly initialized prompts than pre-trained prompts. 

\subsection{Does the number of class affect?}
\label{sec:class_num}
We find that derivative-free prompt learning outperforms derivative-based prompt tuning when the class number of datasets becomes small. In contrast, gradient-based methods achieve better when classifying many categories. We conjecture that this is because datasets with a large number of categories, such as DBPedia (14 classes), can give rise to a higher dimensional search space compared to the counterparts of only two classes, enlarging the range of search space. This still remains an open question for future work.


\end{document}